\documentclass{article}

\PassOptionsToPackage{square, numbers, compress}{natbib}

\usepackage[dblblindworkshop, final]{neurips_2025}
\workshoptitle{AI4Mat}

\usepackage[utf8]{inputenc} 
\usepackage[T1]{fontenc}    
\usepackage{hyperref}       
\usepackage{url}            
\usepackage{booktabs}       
\usepackage{amsfonts}       
\usepackage{nicefrac}       
\usepackage{microtype}      
\usepackage{xcolor}         

\usepackage{bbm} 
\usepackage{multirow}

\usepackage{wrapfig}
\usepackage{graphicx}
\usepackage{amsmath}

\title{One Small Step with Fingerprints, One Giant Leap for \emph{De Novo} Molecule Generation from Mass Spectra}

\author{%
  Neng Kai Nigel Neo \\ \texttt{nnengkai@dso.org.sg} 
      \And
      Lim Jing
      \And
      Ngoui Yong Zhau Preston
      \AND
      Koh Xue Ting Serene
      \And
      Shen Bingquan \\ \texttt{SBingqua@dso.org.sg}
}

\begin{document}

\maketitle

\begin{center}
\vspace{-25px}
DSO National Laboratories \\ Singapore
\end{center}

\begin{abstract}
A common approach to the \emph{de novo} molecular generation problem from mass spectra involves a two-stage pipeline: (1) encoding mass spectra into molecular fingerprints, followed by (2) decoding these fingerprints into molecular structures. In our work, we adopt \textsc{MIST} \cite{MISTgoldmanAnnotatingMetaboliteMass2023} as the encoder and \textsc{MolForge} \cite{ucakReconstructionLosslessMolecular2023} as the decoder, leveraging additional training data to enhance performance. We also threshold the probabilities of each fingerprint bit to focus on the presence of substructures. This results in a tenfold improvement over previous state-of-the-art methods, generating top-1 31\% / top-10 40\% of molecular structures correctly from mass spectra in MassSpecGym \cite{bushuievMassSpecGymBenchmarkDiscovery2024a}. We position this as a strong baseline for future research in \emph{de novo} molecule elucidation from mass spectra.
\end{abstract}


\section{Introduction}

Mass spectrometry (MS) is a foundational technique in analytical chemistry, widely employed for the structural characterisation of small molecules. MS analysis has been used in monitoring reaction intermediates in catalytic cycles and characterising degradation products that govern material longevity. However, the interpretation of mass spectra to elucidate the structure of unknown compounds remains a significant bottleneck, requiring extensive domain knowledge and time-consuming manual effort.
Automated solutions to this problem, known as \emph{de novo} molecule generation from mass spectra, have been proposed over the years. One general approach uses a two-stage pipeline of (1) encoding mass spectra into molecular fingerprints (FP), followed by (2) decoding these fingerprints into molecular structures \citep{bohdeDiffMSDiffusionGeneration2025}. Hence, choosing appropriate encoders and decoders is important. 

Previous approaches utilising this pipeline suffer from poor performance due to the dependency of the fingerprint decoder on the fingerprints generated by the mass spectra encoder. Every bit that the encoder is unsure of leads to compounding errors when using the decoder. 

\begin{figure}
  \vspace{-5px}
  \centering
  \includegraphics[width=0.90\textwidth]{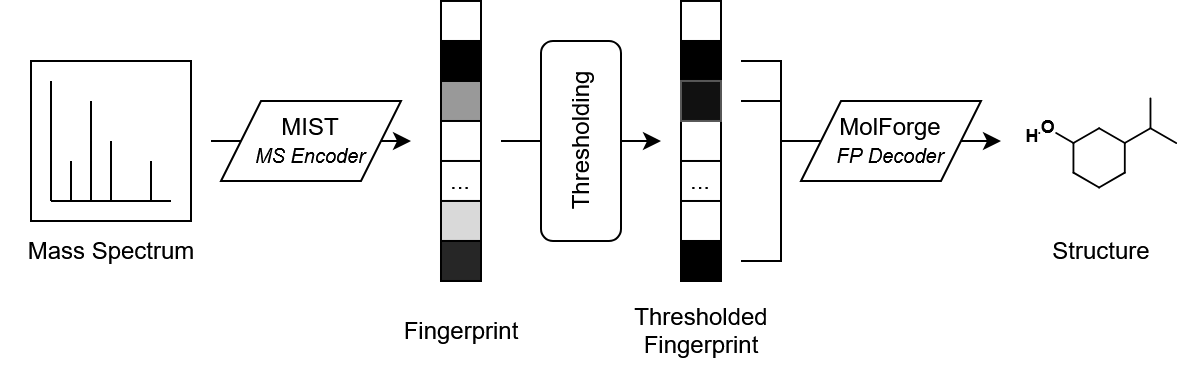}
  \vspace{-10px}
  \caption{\small Our proposed pipeline of using \textsc{MIST} as a mass spectrum encoder, thresholding the fingerprint, and using \textsc{MolForge} as a fingerprint decoder for the \emph{de novo} molecule generation problem.}
  \label{fig:pipeline}
  \vspace{-10px}
\end{figure}

Our core idea lies in choosing the right decoder that mitigates the weaknesses above. Since fingerprints represent the presence and absence of substructures in a molecule, we focus on capturing only present substructures, making it easier for the model to piece together the overall structure. Additionally, a decoder that uses the transformer architecture would be able to scale well with more training data. These two points can be met by using \textsc{MolForge} ~\citep{ucakReconstructionLosslessMolecular2023} as the decoder. Our overall pipeline is shown in Figure \ref{fig:pipeline}. 

Our key contributions are:
\begin{itemize}
    \item We chose \textsc{MIST} \citep{MISTgoldmanAnnotatingMetaboliteMass2023} for spectrum-to-fingerprint encoding with \textsc{MolForge} \citep{ucakReconstructionLosslessMolecular2023} decoder for our two-stage pipeline for \emph{de novo} molecular structure generation from mass spectra.
    \item We demonstrate that training the fingerprint-to-structure decoder on a larger dataset with varying chemical domains significantly enhances generalisation and structure recovery, even when fingerprint similarity to ground truth is moderate.
    \item Our method outperforms previous state-of-the-art by an order of magnitude, achieving 31\% top-1 and 40\% top-10 structure generation accuracy on the MassSpecGym \citep{bushuievMassSpecGymBenchmarkDiscovery2024a} dataset. 
\end{itemize}

\section{Problem Statement \& Definitions}
\label{ch:problem}

We adopt the problem definition and metrics (top-$k$ accuracy, Maximum Common Edge Subgraph (MCES) and Tanimoto similarity) for \emph{de novo} molecule generation from MassSpecGym \citep{bushuievMassSpecGymBenchmarkDiscovery2024a}. Further details can be found in Appendix \ref{app:definitions}. Importantly, as this problem is \emph{de novo} in nature, algorithms attempting to solve this problem need to generate molecular structures without any reference libraries. 

\section{Methodology}
\label{ch:methodology}

Our current method utilises a combination of existing models. We show how we have adapted these models to form a pipeline to solve the \emph{de novo} molecule generation problem. These models (\textsc{MIST}, \textsc{MolForge}) have open source implementations available. Details can be found in Appendix \ref{app:implementation}. 

\begin{wrapfigure}{r}{0.5\textwidth}
  \vspace{-40px}
  \centering
  \includegraphics[width=0.49\textwidth]{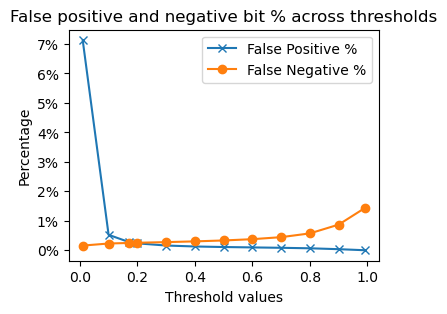}
  \caption{\small False positive and negative bits across different threshold values when applied to MIST fingerprints.}
  \label{fig:fpfn}
  \vspace{-20px}
\end{wrapfigure}

\subsection{Mass Spectra Encoder}

We use the pretrained \textsc{MIST} models from \textsc{DiffMS}. The \textsc{MIST} model was trained on MassSpecGym to predict molecular fingerprints from mass spectra, where each peak is annotated with possible chemical formulae by an automated annotation program. We refer the reader to Section 3.3 of the \textsc{DiffMS} paper for technical details. In short, each peak in the mass spectra is encoded via a chemical formula transformer, then passed to a module that outputs the predicted fingerprint. 

\subsection{Fingerprint Decoder}

\textsc{MolForge} \citep{ucakReconstructionLosslessMolecular2023} is an autoregressive transformer model that predicts molecular structure from fingerprints. Specifically, it takes in the on-bits of a fingerprint (bits that have a value of `1') to predict a SMILES string that represents the molecular structure. This is done using an encoder-decoder model that generates a mapping of indices for the encoder and a mapping of tokens (representing atoms, parenthesis, and other symbols in SMILES strings) for the decoder.

\textsc{MolForge} takes the indices of on-bits of the molecular fingerprint as input. However, \textsc{MIST} generates fingerprints as probabilities, real values between 0 to 1. As such, we set a threshold $t$ and use all indices of bits $\geq t$ as input to \textsc{MolForge}. 
We investigate two different threshold values: a fixed value $t=0.5$, and in Section \ref{sec:prior-adjusted} a prior-adjusted threshold which assumes that the distribution of on-bits in the training set is similar to that of the test set. Specifically, we choose $t$ such that the proportion of logits thresholded to 1 is equal to the overall proportion of on-bits in the ground truth fingerprints of the training set, and we then apply this threshold to the test set logits. As shown in Figure \ref{fig:fpfn}, this prior-adjusted threshold can be calculated without any data leakage from the test set, and still results in a good balance of false positive and negative bits when compared to the ground truth fingerprints.

MassSpecGym only contains a limited number of molecules in the dataset ($\sim$17k), as there are multiple mass spectra for the same molecule. Thus, there is a need for further training such that \textsc{MolForge} has more data to learn the decoding process. We train \textsc{MolForge} with the same dataset used in \textsc{DiffMS} and its baselines, which contains $\sim$3M compounds and does not include any molecules in the test set. \textsc{MolForge} was trained to take in 4096-bit fingerprints as the pretrained \textsc{MIST} models generate fingerprints with a length of 4096 bits. 

To generate the top-$k$ molecular predictions, \textsc{MolForge} employs a beam search decoding strategy. At each decoding step, the model expands the most probable partial SMILES sequences by selecting the top-ranked token continuations according to the cumulative log-probability of each sequence. The search proceeds iteratively until complete SMILES strings are formed, after which the top-$k$ highest probability sequences are retained as final outputs. 

\subsection{Datasets}

We use MassSpecGym \citep{bushuievMassSpecGymBenchmarkDiscovery2024a} to test our pipeline. For training on additional compounds, we refer to the combined dataset that is made available in \textsc{DiffMS} \citep{bohdeDiffMSDiffusionGeneration2025}, which corresponds to data from DSSTox \citep{ccteDistributedStructureSearchableToxicity2024}, HMDB \citep{wishartHMDB50Human2022}, COCONUT \citep{sorokinaCOCONUTOnlineCollection2021}, and MOSES \citep{polykovskiyMolecularSetsMOSES2020} datasets. This results in about 2.8 million unique compounds, and we use this dataset as additional training for \textsc{MolForge}. 

\section{Results and Discussion}

\begin{table*}[t]
\caption{\textit{De novo} structural elucidation performance on MassSpecGym~\citep{bushuievMassSpecGymBenchmarkDiscovery2024a} dataset, across accuracy (Acc), MCES edit distance (MCES) and Tanimoto similarity of fingerprints (Tanimoto). For our pipeline (MIST+MolForge), we used a fixed threshold $t$ of 0.5 and a prior-adjusted threshold value of 0.172, indicated by *. 
}
\vspace{-0.05in}
\label{table:main_results}
\begin{center}
{
\small
\begin{tabular}{lcccccc}
\toprule
& \multicolumn{3}{c}{Top-1} & \multicolumn{3}{c}{Top-10} \\
\cmidrule(lr){2-4}
\cmidrule(lr){5-7}
Model & Acc. $\uparrow$ & MCES $\downarrow$ & Tanimoto $\uparrow$ & Acc. $\uparrow$ & MCES $\downarrow$ & Tanimoto $\uparrow$ \\
\midrule
\midrule
MADGEN & {1.31\%} & 27.47 & {0.20} & {1.54}\% & {16.84} & {0.26} \\
DiffMS & 2.30\% & 18.45 & 0.28 & 4.25\% & 14.73 & 0.39 \\
MIST+MolForge, t=0.5 & 28.27\% & 14.72 & 0.64 & 36.11\% & 10.69 & 0.70 \\
\textbf{*MIST+MolForge, t=0.172} & \textbf{30.97\%} & \textbf{12.38} & \textbf{0.68} & \textbf{40.04\%} & \textbf{8.63} & \textbf{0.74} \\
\bottomrule
\end{tabular}
}
\end{center}
\vskip -0.1in
\end{table*}

\subsection{\emph{De novo} molecule generation}

The primary results of our model pipeline on the \emph{de novo} molecular generation task are shown in Table \ref{table:main_results}. We benchmark our pipeline against the state-of-the-art \textsc{DiffMS} model, along with other established baselines reported in prior work. Our approach yields a significant performance gain, achieving an approximately 10-fold increase in exact structure match accuracy when \textsc{MolForge} is used as the fingerprint-to-structure decoder. Furthermore, the Tanimoto similarity between the predicted and ground truth molecular fingerprints show good agreement. The Maximum Common Edge Substructure (MCES) metric also shows a reduction in the number of extra atoms required to form a common graph between predicted and reference molecules, implying that predicted structures are more similar to the ground truth structure compared to other methods.

\subsection{Performance of \textsc{MIST} as mass spectra encoder}
As we have utilised the same pretrained \textsc{MIST} model from the \textsc{DiffMS} paper, we refer the reader to Section 4.4 of the \textsc{DiffMS} paper. \textsc{DiffMS} has also conducted an ablation study, showing that pretraining \textsc{MIST} results in improved performance compared to without pretraining \textsc{MIST}. 

\textsc{MIST} also takes in annotations, in terms of chemical formulae, of peaks in the mass spectrum.  
In our study, removing annotations from the mass spectra (i.e. not supplying chemical formulae of the peaks into \textsc{MIST}) results in a decrease in Tanimoto Similarity (as calculated between fingerprint of the ground truth molecule and fingerprint predicted by MIST) from 0.731 to 0.627. Hence, peak annotations of the mass spectra are also essential for the good performance of our proposed pipeline. 

\subsection{Performance of MolForge as fingerprint decoder}

\begin{table*}[t]
\caption{Performance of \textsc{MolForge} on different training datasets and the input fingerprint (FP) used. Additional training data leads to improved performance for \textsc{MolForge}. We also compare ground truth fingerprints, generated directly from the ground truth structure, with fingerprints from MIST ($t=0.172$), which are predicted from the corresponding mass spectra.}
\vspace{-0.05in}
\label{table:molforge}
\begin{center}
{
\begin{tabular}{ll|cccc}
\toprule
& & \multicolumn{2}{c}{Top-1} & \multicolumn{2}{c}{Top-10} \\
\cmidrule(lr){3-4}
\cmidrule(lr){5-6}
MolForge training dataset & Input FP & Acc. $\uparrow$ & Tanimoto $\uparrow$ & Acc. $\uparrow$ & Tanimoto $\uparrow$ \\
\midrule
\midrule

Combined DiffMS ($\sim$3M) & From MIST & 30.97\% & 0.68 & 40.04\% & 0.74 \\
Combined DiffMS ($\sim$3M) & Ground truth & 46.00\% & 0.89 & 59.28\% & 0.93 \\
MassSpecGym ($\sim$17k) & Ground truth & 0.00\% & 0.18 & 0.00\%  & 0.19 \\

\bottomrule
\end{tabular}
}
\end{center}
\vskip -0.5cm
\end{table*}

We trained \textsc{MolForge} with the same dataset used in \textsc{DiffMS} and its baselines, which contains more than two million compounds. This allowed it to better capture the distribution of SMILES strings compared to just using the MassSpecGym dataset, which has only $\sim$17k compounds. Without these additional data, \textsc{MolForge} fails to generate any structures from the test set correctly (see Table \ref{table:molforge}). The increase in performance with larger dataset size is expected due to data scaling laws \citep{kaplanScalingLawsNeural2020}. Hence, we consider this step to be vital. 

It is also important to note that the performance of \textsc{MolForge} with ground truth fingerprints as input is very high with 59\% top-10 accuracy (in terms of molecule structural matches) in the test set. This shows that \textsc{MolForge} is a good model for the inverse problem of decoding fingerprints to structure, and why it is a vital component of our pipeline. 

\begin{figure}
  \centering
  \includegraphics[width=0.85\textwidth]{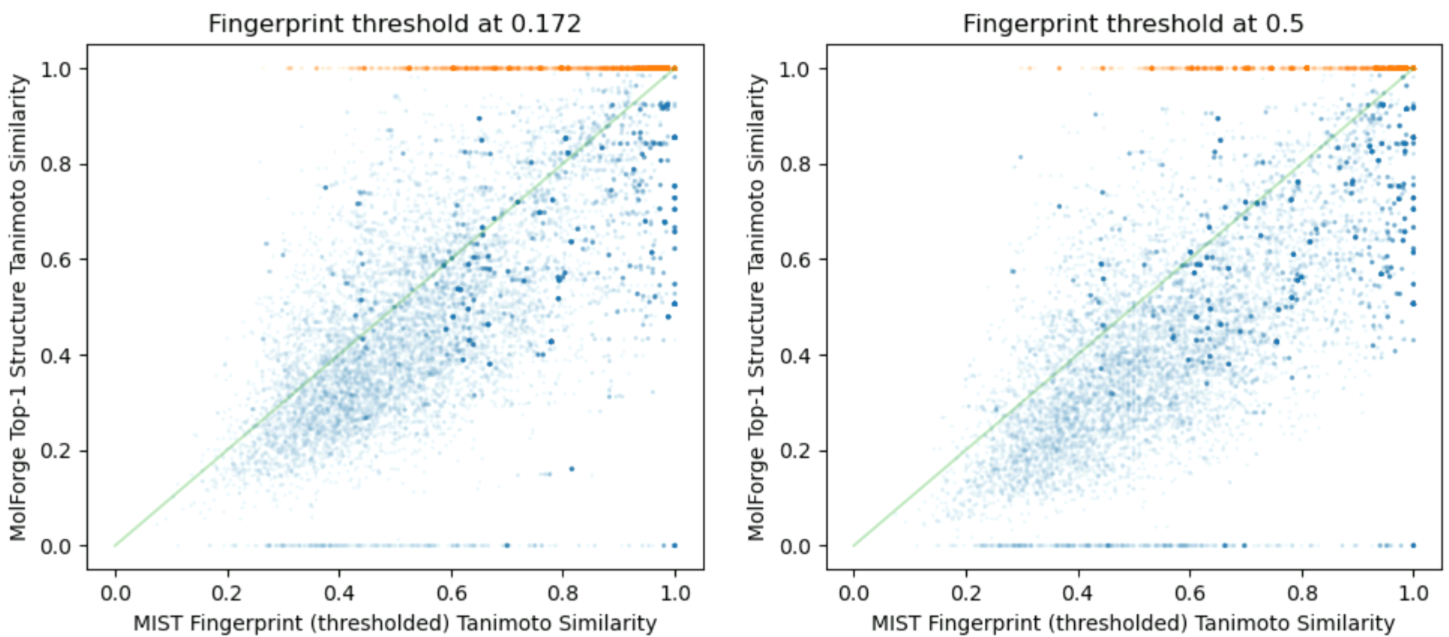}
  \caption{\small Tanimoto Similarity of thresholded \textsc{MIST} fingerprint and Top-1 structure from \textsc{MolForge}, both compared to the ground truth fingerprint. Orange points indicate an exact structural match, and points above (below) the green parity line show predicted structures that are better (worse) than the thresholded fingerprint.}
  \label{fig:tsvsts}
  \vspace{-10px}
\end{figure}

Additionally, instead of passing the probability of each bit in the fingerprint, thresholding the probabilities as a step function helps focus the decoder on the presence of substructures. \textsc{MolForge} only takes in the indices of the fingerprint that has a value of 1, representing the presence of probable substructures in the fingerprint. 
This makes it capable of recovering accurate molecular structures even when the fingerprints predicted by \textsc{MIST} only moderately resembles the ground truth in terms of Tanimoto similarity, as shown in Figure \ref{fig:tsvsts}. \textsc{MolForge} predicts structures that has a better match to the ground truth in terms of Tanimoto similarity, compared to the thresholded fingerprint it was given as input, at 27\% of the test set for $t=0.5$ and 36\% for a prior-adjusted threshold of $t=0.172$.

Hence, the elevated performance of \textsc{MolForge} with ground truth fingerprints shows that the key bottleneck in our pipeline is the prediction of fingerprints from mass spectra. The performance of our pipeline is close to the best-case scenario if \textsc{MIST} were to predict fingerprints perfectly, suggesting room for improvement. Even though the fingerprint predicted by \textsc{MIST} has errors, \textsc{MolForge} is able to overcome some of them and generate the correct structure. Further improvements on the prediction of fingerprints from mass spectra should translate well to better performance in the \emph{de novo} molecule generation problem given the well-rounded performance of \textsc{MolForge}. 

\subsection{Investigation on using prior-adjusted thresholds}
\label{sec:prior-adjusted}
The prior-adjusted threshold method calculates the prior distribution of on-bits in the training set and selects a threshold such that the proportion of thresholded logits matches this prior. For the MassSpecGym training set, the proportion of on-bits is 1.09\%. To preserve this proportion, we determine the logit value corresponding to the top 1.09\% of logits, which equals 0.172. Accordingly, we set the threshold to $t=0.172$. 

As shown in Table \ref{table:main_results}, using this prior-adjusted threshold results in a slight boost in performance, with the top-1 accuracy of 31\% and top-10 accuracy of 40\%. This is mainly attributed to the better balance of false positives and false negatives in the thresholded fingerprint. Hence, the indices passed to \textsc{MolForge} would be more similar to the ground truth which results in the increased performance. We recommend using the prior-adjusted threshold value when using this pipeline.  

\section{Relevant Literature}
\label{ch:literature}

\textbf{\emph{De novo} molecule generation from mass spectra}. 
The problem of \emph{de novo} molecular generation from mass spectrometry data, also referred to as the inverse problem of structural elucidation from mass spectra, has garnered significant attention in computational chemistry and machine learning. Early approaches leveraged machine learning techniques to map mass spectra to molecular structures \citep{limChemicalStructureElucidation2018, weiRapidPredictionElectron2019}. 
The MassSpecGym benchmark \citep{bushuievMassSpecGymBenchmarkDiscovery2024a} represents one of the most comprehensive open-source datasets for tandem mass spectrometry-based structure elucidation to compare across models. 
More recently, \textsc{DiffMS} \citep{bohdeDiffMSDiffusionGeneration2025}, a generative diffusion model, has been proposed to solve this problem. The method first employs \textsc{MIST} \citep{MISTgoldmanAnnotatingMetaboliteMass2023} to predict molecular fingerprints from mass spectra. These fingerprints are then used as conditional inputs to a graph diffusion model that generates candidate molecular structures as graphs. This two-stage framework enables \textsc{DiffMS} to achieve a test-set accuracy of 2.30\% on the MassSpecGym benchmark.

\textbf{Encoding mass spectra to molecular fingerprints}. DeepEI \citep{DeepEIjiPredictingMolecularFingerprint2020} also encodes a mass spectrum, but does not fully generate a molecular fingerprint. 
CSI:FingerID \citep{CSIduhrkopSearchingMolecularStructure2015}  uses a fragmentation tree to generate plausible fragments. However, this fingerprint type is different from Morgan fingerprints, and thus is not as extensible for usage with other fingerprint decoder models. 

\textbf{Decoding molecular fingerprints to molecular structure}.
\textsc{DiffMS} \citep{bohdeDiffMSDiffusionGeneration2025}, as stated earlier, contains a conditional graph diffusion model that takes in a fingerprint representation to predict molecule structure. 
Other fingerprint-to-structure models can be paired with \textsc{MIST}. Neuraldecipher \citep{leNeuraldecipherReverseengineeringExtendedconnectivity2020} uses a standard feed-forward model to decode the molecular fingerprint, while MSNovelist \citep{stravsMSNovelistNovoStructure2022} uses a recurrent neural network with long short-term memory architecture. 

\section{Conclusion}

Various two-stage pipelines have been proposed for solving the \emph{de novo} molecular generation problem from mass spectra, comprising (1) encoding mass spectra into molecular fingerprints, followed by (2) decoding these fingerprints into molecular structures. Leveraging external chemical datasets significantly enhances performance in both stages, especially so for the decoder. This stems from the limited availability of labeled mass spectra, in contrast to the abundance of molecular fingerprint data that can be exploited during decoder training. Notably, substituting the decoder with \textsc{MolForge}, a seemingly modest architectural change, yields a tenfold improvement over prior state-of-the-art, highlighting the how choosing the right decoder is critical. 
\textsc{MolForge} uses on-bit indices of the fingerprint as input, and we show how using a prior-adjusted threshold leads to further improvement compared to a predetermined value.
We position this pipeline as a strong baseline for future work in \emph{de novo} structure prediction from mass spectra, and highlight the spectra-to-fingerprint encoding step as a promising direction for further investigation.

\newpage

\bibliography{neurips_2025}
\bibliographystyle{dinat}

\medskip


\appendix

\section{Problem Statement and Definitions}
\label{app:definitions}

We adopt the problem definition for \emph{de novo} molecule generation from MassSpecGym \citep{bushuievMassSpecGymBenchmarkDiscovery2024a}, which involves predicting a molecule’s structure $G$ given its corresponding mass spectrum $MS$. The molecule is modeled as a graph $G = (V, E)$, where the $N$ atoms are represented by the set of vertices $V \in \mathbb{V}^N$, and the $M$ chemical bonds are represented by the set of edges $E \in \mathbb{E}^M$. The mass spectrum $MS$ consists of intensity values $Y \in (0,1]$ associated with mass-to-charge ratios $X \in \mathbb{R}_+$.

This paper focuses on \emph{de novo} molecule generation with the chemical formula provided as an additional input, which is also part of the MassSpecGym benchmark. In this setting, the chemical formula, which corresponds to the atom set $V$, is known in advance. The task is to predict the molecular structure $G$ given its mass spectrum $MS$ and chemical formula. 

Importantly, as this problem is \emph{de novo} in nature, algorithms attempting to solve this problem need to generate molecular structures without any reference libraries. 

\subsection{Metrics}

\textbf{Top-$k$ metrics.} The \emph{de novo} molecule generation problem can be formulated as predicting a set of $k$ candidate graphs $\widehat{G}_k = \{\widehat{G}_1, \ldots, \widehat{G}_k\}$, rather than a single predicted graph $\widehat{G}$. Especially since mass spectra may not contain enough information to predict structures accurately, this formulation better reflects the uncertainty and complexity of molecule generation from mass spectra.

To evaluate the quality of the predicted molecular graphs $\widehat{G}_k$, we compare them against the ground-truth graph $G$ using three metrics, following MassSpecGym \citep{bushuievMassSpecGymBenchmarkDiscovery2024a}. As a summary:

\begin{enumerate}
    \item \textbf{Top-$k$ accuracy.} 
    \begin{equation}
        \text{Top-}k\ \text{accuracy:}\ \mathbbm{1}\{G \in \widehat{G}_k\}
    \end{equation}
is the presence of the ground truth molecule within the top-$k$ predictions of the model. This is averaged over all test examples. Here, $\mathbbm{1}\{\cdot\}$ is the indicator function, which returns 1 if the condition is true and 0 otherwise. 
    \item \textbf{Maximum Common Edge Subgraph (MCES) metric.}~\citep{MCESkretschmerCoverageBiasSmall2025}
    \begin{equation}
    \text{Top-}k\ \text{MCES:}\ \min_{\widehat{G} \in \widehat{G}_k} \text{MCES}(G, \widehat{G}), \label{eq:topk-mces}
    \end{equation}
The MCES metric measures the graph edit distance, or the number of edges that need to be added to candidate structure $\widehat{G}$ such that ground truth $G$ is also a subgraph of it. We report the best similarity score among the top-$k$ candidates as averaged across the test set. A score of 0 indicates identical graphs, while larger values correspond to greater structural dissimilarity.

\item \textbf{Tanimoto similarity}, computed on the Morgan fingerprints of the molecules \citep{morganGenerationUniqueMachine1965}. 
\begin{equation}
\text{Top-}k\ \text{Tanimoto:}\ \max_{\widehat{G} \in \widehat{G}_k} \text{Tanimoto}(G, \widehat{G}). \label{eq:topk-tanimoto}
\end{equation}
This score reflects how well the generated molecule captures true molecular substructures. Tanimoto similarity ranges from 0 to 1, with 1 indicating perfect structural similarity (but not exact similarity, due to the possibility of fingerprint collisions).

\end{enumerate}

\subsection{Cheminformatics terms}

While exact graph matching can be used to check if the ground truth and predicted molecules match, the cheminformatics domain has also developed other representations for this. In this paper, we focus on using InChIs \citep{hellerInChIIUPACInternational2015} as textual representations of molecules. InChIs are guaranteed to be unique for each structure and thus can be used as a substitute to test for molecule similarity. SMILES strings \citep{weiningerSMILESChemicalLanguage1988}, though a common text representation of molecules, are not used for comparison, as multiple SMILES strings can be used to refer to the same molecule.  

We also use molecular fingerprints as a representation of a molecule, which are usually 2048 or 4096 bits in length. Molecular fingerprints are bit vectors that represent the presence (`1' / on-bit) or absence (`0' / off-bit) of substructures within a molecule. One of the most commonly used fingerprints is a Morgan fingerprint \citep{morganGenerationUniqueMachine1965}. 

\section{Implementation Details}
\label{app:implementation}

Open source implementations are available for the models (\textsc{MIST} \footnote{https://github.com/samgoldman97/mist/tree/main\_v2}, \textsc{MolForge} \footnote{https://github.com/knu-lcbc/MolForge}) in our pipeline. We have generally followed the default settings for these models, and further modifications are explained in this section. 

\textbf{MIST}: We use pretrained \textsc{MIST} models from \textsc{DiffMS} \footnote{https://zenodo.org/records/15122968}, and base our implementations from the \textsc{DiffMS} code and annotated MassSpecGym data. We proceed with the inference run with the \textsc{MIST} models pretrained for MassSpecGym, and collect the \texttt{output} of the \texttt{encoder} (\textsc{MIST} model) in \texttt{test\_step} as the fingerprint outputs. We extracted all indices for which its corresponding bit is at least the threshold value. 

\textbf{MolForge}: The training data was formatted following the examples in the \textsc{MolForge} Github repository, in which SMILES strings were split into tokens (i.e. atoms, brackets or numbers) and fingerprints were generated using the Morgan fingerprint generator from \textsc{RDKit} \footnote{https://www.rdkit.org} with a radius of 2 and fingerprint size of 4096 bits. We first train the tokenizer for this model with the additional training data as provided in \textsc{DiffMS}. Following which, we trained the model on the same dataset with a learning rate of 5e-4 and batch size of 128 for 6 epochs. This took about 3 days to run with a Nvidia A40 GPU. For inference, we used the beam search code from \textsc{MolForge} with a beam size of 10.

\end{document}